# An analysis of the combination of feature selection and machine learning methods for an accurate and timely detection of lung cancer

Omid Shahriyar, Babak Nuri Moghaddam, Davoud Yousefi, Abbas Mirzaei [1], Farnaz Hoseini[2]

1. Department of Computer Engineering, Ardabil Branch, Islamic Azad University, Ardabil, Iran
2. Department of Computer Engineering, National University of Skills (NUS), Tehran, Iran

**Article Info**

**Article History:**
*Received*
*Revised*
*Accepted*

*DOI:*

*Corresponding Author's Email Address:*
mirzaei_class_87@yahoo.com

**Abstract**

One of the deadliest cancers, lung cancer necessitates an early and precise diagnosis. Because patients have a better chance of recovering, early identification of lung cancer is crucial. This review looks at how to diagnose lung cancer using sophisticated machine learning techniques like Random Forest (RF) and Support Vector Machine (SVM). The Chi-squared test is one feature selection strategy that has been successfully applied to find related features and enhance model performance. The findings demonstrate that these techniques can improve detection efficiency and accuracy while also assisting in runtime reduction. This study produces recommendations for further research as well as ideas to enhance diagnostic techniques. In order to improve healthcare and create automated methods for detecting lung cancer, this research is a critical first step.

## 1. Introduction

One of the main causes of cancer-related deaths globally is thought to be lung cancer. The disease's five-year survival rate is still low, even with major advancements in treatment. The primary causes of this are delayed diagnosis and the shortcomings of conventional diagnostic techniques. Thus, it is especially crucial to develop novel and efficient techniques for lung cancer early detection [1].

New approaches to improving the diagnostic process have been made possible by recent advancements in data science and machine learning. Machine learning algorithms like Support Vector Machine (SVM) and Random Forest (RF) have proven to be quite effective in analyzing complex and vast amounts of data. These algorithms improve the precision and effectiveness of diagnostic models by employing feature selection techniques [2].

One of the most important phases in the machine learning process is feature selection, which helps to improve the accuracy of prediction models while simultaneously lowering computing complexity by minimizing the dimensions of the input. The Chi-squared test is an important statistical technique for determining related traits and eliminating unrelated ones. This approach has proven to be successful in analyzing medical data, particularly when it comes to lung cancer diagnosis [3-4].

In this review paper, current developments in the use of machine learning algorithms to the detection of lung cancer are thoroughly and methodically reviewed. To provide the best answers for further research, we aim to assess and contrast various feature selection strategies and classification algorithms, pointing out the advantages and disadvantages of each. The findings of this study can assist researchers and experts in creating diagnostic instruments that are quicker and more accurate [5].

### Materials and methods

### 1. The chi-square test

We first looked at publications on the diagnosis of lung cancer before writing this section of the review study. Genetic algorithms have been used as a transcendence technique in the majority of earlier research. However, because of the numerous iterations and fitness value computation, this technique takes a long time to produce results. In order to identify traits linked to the diagnosis of lung cancer, recent research has resorted to statistical techniques like the Chi-square test. According to the findings, the Chi-squared test is more successful and efficient at predicting the diagnosis of cancer [6-7].

### 1.1. Target

In order to diagnose lung cancer, this study compares the effectiveness of two models: Random Forest (RF) and Support Vector Machine (SVM). The performance of models with and without the Chi-squared feature selection technique has been analyzed in this context, and related features have been found using this technique. The study's ultimate objective is to shorten the model's implementation time while improving the model's accuracy, recall rate, and correctness.[8]

### 1.2. Suggested methods
#### 1.2.1. Feature Selection

Cutting down on superfluous features is a crucial step in enhancing machine learning models' performance. Important characteristics were chosen for the investigation using the Chi-squared statistical test. This test looks at the connection between the intended outcome and batch characteristics. The Chi-squared value is the lowest when two qualities are independent. Conversely, a feature is more dependent on the response and

The flowchart of the current work is displayed in Figure 1. Outlining the various sections and subsections aids in a better understanding of the article and helps to solidify its structure in the reader's mind.

more suited for model training if its Chi-squared value is higher.[9]

#### 1.2.2. Support vector machine (SVM)
One of the best methods for predicting lung cancer is SVM. This algorithm establishes a boundary between two data classes in order to teach the model. The SVM model for both linear and nonlinear data was investigated in this study. A kernel function that transforms data into linear space is utilized for nonlinear data.[10]

#### 1.2.3. Random Forest:

RF is a classification technique that makes predictions using a collection of decision trees. In this work, lung cancer data was categorized using RF, and the outcomes were compared with SVM.[11]

### 1.3. Explanation of the data

The data utilized in this study was taken from the data site. world and included 25 features and 1000 samples. This dataset's objective is to categorize lung cancer into three groups: low, medium, and high. Subsequently, duplicate and null values were eliminated from the data using pre-processing. After that, the data were split into two sets: experimental (35%) and training (65%).[12]

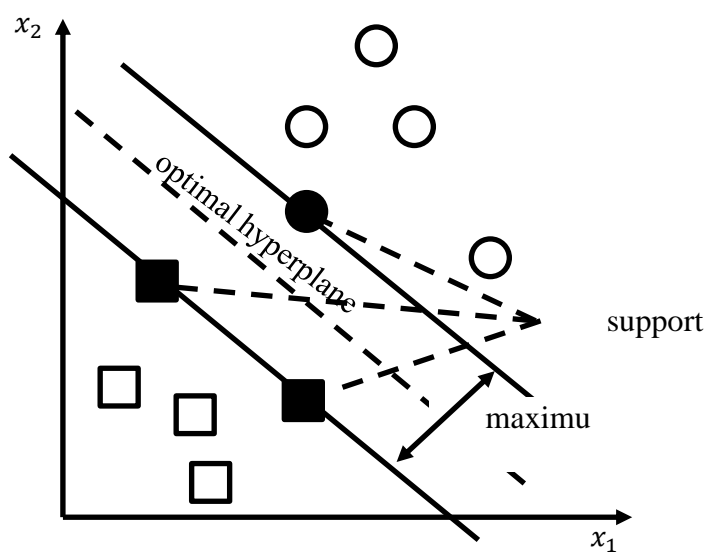

Figure 1: Support vector machine

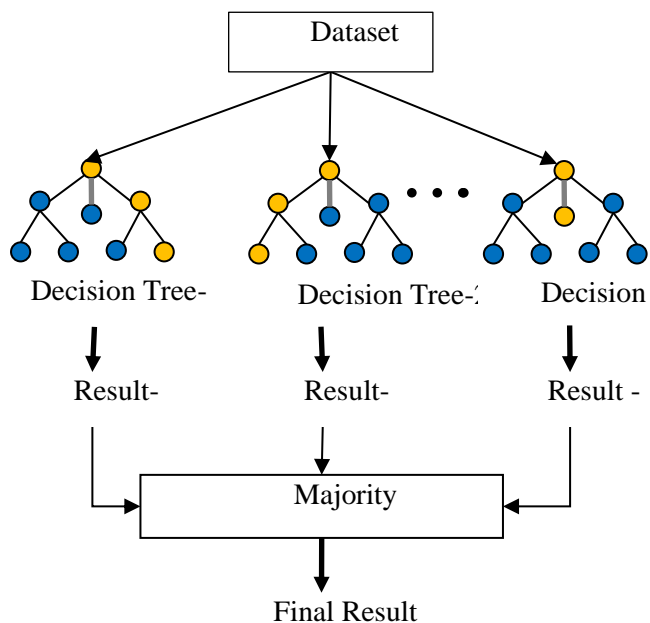

Figure 2: Random Forest

The data was then subjected to the Chi-squared feature selection technique, which resulted in the selection of the best features associated with the target. Ultimately, data was classified using two SVM and RF classification algorithms, and the outcomes were compared according to runtime, accuracy, recall rate, and accuracy [13].

### 1.4. Proposed methodology

In this Section, various processes are listed, from collecting data to processing them and completing various steps to achieve the desired results. Data is collected and reviewed first. Then the data preprocessing step is done to remove empty and duplicate values. After that, the data is divided into educational and experimental sets in a ratio of 65% and 35%. Finally, three classification algorithms, random forest and support vector machine are applied to data classification and the comparison of these two algorithms is done with and without the use of the Chi-squared feature selection [14-15].

### 2. Microarray data

Gene expression values as features have been the focus of this study's usage of lung cancer microarray data. The Kent Ridge Biomedical Dataset database at the University of Michigan is the source of this dataset, which may be accessed at http://datam.i2r.astar.edu.sg/datasets/krbd. Two classes were created from the 7,129 features in the lung cancer dataset: 86 samples with cancer and 10 samples without disease [16].

### 2.1. Feature selection based on the kernel function

In this investigation, attributes were chosen using the Cornell function, which was first presented by Vapnik [7] and later refined by Scholkopf [8] and his associates as well as Christianini and Taylor [9]. The kernel function transfers linear and nonlinear problems to a higher-dimensional space and solves them using a linear classification [17]. The following is the expression for the kernel function:

$$K(X_i, X_j) = \Phi(X_i).\Phi(X_j) \quad (1)$$

In this study, Gaussian kernel was used to select properties that are used to measure the inconsistency between properties. The inconsistency between properties is calculated using the Gaussian kernel through the following equation:

$$K(X_i, X_j) = exp\left(-\frac{\|X_i - X_j\|^2}{2\sigma^2}\right) \qquad (2)$$

Objective the objective function in this feature selection method is to optimize the weight of properties by minimizing inconsistencies between samples and the respective cluster centers. The weight updating process is done repeatedly until the convergence is achieved.[18]

### 2.2. Feature selection based on SVM-RFE

SVM-RFE is another feature selection technique used in the study, introduced in 2002 by Gayon. SVM-RFE is an embedded method that uses a weight vector obtained from SVM training to rank features. In each iteration, properties that have the lowest rating criteria are omitted. The ranking score is determined using weighted vector components as follows:

$$R = w_j^2 \qquad (3)$$

### 2.3. Backup vector machine (SVM)

The backup vector machine (SVM), developed in 1992 by Bosser, guyon and vapnik, was used in this study to classify lung cancer data. The main objective of SVM is to find an optimal superpage that will give the maximum margin between the two classes [23]. The SVM optimization problem is formulated as follows:

$$minimize \quad \frac{1}{2}\|w\|^2 + C\sum_{i=1}^{N}\xi_i \qquad (4)$$

$$subject\ to \quad Y_i(w.X_i + b) \geq 1 - \xi_i, \xi_i \geq 0$$

Where C controls the balance between maximizing margin and reducing classification error.

### 2.4. Performance criteria

The performance of the classification model is evaluated using standard criteria such as accuracy, positive accuracy, recall and F1 rating. These metrics are calculated based on the confusion matrix, which shows the number of true positive (TP), true negative (TN), false positive (FP) and false negative (FN) cases [24].

Table 1. Confusion matrix

| Predicted class | Real class 1 | Real class 2 |
|---|---|---|
| class 1 | TP | FN |
| class 2 | FP | TN |

*TP (true positive) the number of Class 1 experimental data predicted as Class 1.

Table 2. A sample of lung cancer

| | A28102_at(Nucleotide) | AB000114_at(Osteomodulin) | … | Z97074_at(RABEPK) |
|---|---|---|---|---|
| NO | 1 | 2 | ….. | 7129 |
| X1 | 170 | 69.4 | ….. | 276 |
| X2 | 59.7 | 18.1 | ….. | 134.7 |
| …… | …… | ….. | ….. | ….. |
| X86 | 57 | 7.9 | ….. | 50.7 |
| …… | …… | ….. | ..... | …… |
| X96 | 106.2 | 164 | ….. | 140.5 |

TN (true negative): the number of Class 2 experimental data predicted as Class 2.
FP (false positive): the number of Class 2 experimental data predicted as Class 1.
FN (false negative): the number of Class 1 experimental data predicted as Class 2

*Note: A28102, AB000114_at and Z97074_at show the data properties.
The numbers in each row represent the expression of a gene in all experiments.
X1 ،X2 ،... X96 shows samples.

## 3. Random release (SDS)

In this study, symbolic data with radiomic properties was used. In this section, gray-level co-occurrence matrices (GLCM) and the Gabor filter feature extraction method were used, as well as the SDS feature selection method using decision tree classifications, neural network (NN) and Naïve Baye [25].

### 3.1. Dataset

The study used the Atlas genome cancer data set (TCGA), which includes 140 normal and 130 abnormal images. For symbolic data, lung histology has been extracted in up to three different time frames [26].

### 3.2. Extraction of radiomic properties

The process of gathering texture, shape, and color information is called feature extraction. Features include search, retrieval, and storing of pertinent information utilized in image processing. Reducing the resources required to describe large amounts of data is the goal of feature extraction. Texture and form are examples of quantitative attributes that are employed in the extraction of radiomic properties from photographs. In this approach, tissue property vectors were extracted using gray-level co-occurrence matrices (GLCM), and shaped properties were extracted using the Gabor filter [27-28]

Haralick's 1973 grey surface coincidence matrix (GLCM) is still a widely used tissue examination technique. By determining the frequency of pixel pairings with specific values in a spatial connection inside the image, GLCM is able to identify tissue. Identification of the object in the desired area (ROI) is aided by tissue features [29-31].

Gabor filters are linear filters defined using the harmonic function and the gossip function. The formula for the Gabor filter is as follows:

$$G_c[i,j] = Be^{\frac{i^2+j^2}{2a^2}} \cos\left(2\prod f(i\cos\theta + j\sin\theta)\right) \quad (5)$$

$$G_s[i,j] = Ce^{\frac{i^2+j^2}{2a^2}} \sin\left(2\prod f(i\cos\theta + j\sin\theta)\right) \quad (6)$$

Where B and C are normalization factors that need to be determined, and σ specifies the filter effect Area size and θ the Gabor filter angle to adjust the size.

### 3.3. Feature selection based on random release search algorithm (SDS)

The SDS algorithm for selecting properties through communication is used to evaluate the subset of properties in an effective way. In the early stages, each factor is assigned to combine the attribute subset of its search space. Each factor uses an independent, random division of the data set to form training and testing subgroups with ratios of 80% and 20%. The hypothesis here is a binary string representing a subset of properties. If the bit is equal to 1, the corresponding property is included, and if it is 0, it is not included.[32-34]

A classification is used to determine the operator actions based on the life function during the test phase, and the predictive accuracy is compared [35]. The release phase starts once this procedure is repeated for the operators to ascertain their status. By choosing the active agent, each passive agent passes on its notion to the passive agent during the release phase. If not, a new hypothesis is chosen from the search space by the factors that were chosen. The following is the SDS formula used to choose the feature [36-37]:

**Primary phase**
Allocation of factors to random assumptions with passive states
**Evaluation phase**
- Assessing the amount of life.
- Finding the best amount of life.
**Testing phase**
- If the operator's life is greater than the random operator's life, the operator is activated; otherwise, it is deactivated.

**Release phase**

- If the agent is disabled, selecting a random agent
- If the selective factor is active, it copies and changes its hypothesis and evaluates the value of life.

-Otherwise, it selects a new hypothesis from the search space and evaluates the value of life.

This test and release cycle continues to the number of permissible iterations to achieve the optimal feature subset.

### 3.4. Classifications used
### 3.4.1. Decision tree

Decision tree is used as a predictive model to map observations to conclusions about the target value. The decision tree algorithm recursively divides the dataset using deep-digging approaches from time to time so that all data belongs to a specific class. The structure of the decision tree consists of roots, internal nodes and leaf nodes. Each internal node indicates the test condition on the property, and each branch displays the test result, and each leaf node is assigned to a class tag [38-40].

### 3.4.2. Naïve Bayes

The Naïve Bayes classification is based on the Bayes theorem and is suitable for data with high input dimensions. This method can simply perform well in complex situations and, using supervised learning, predicts the chances of belonging to different classes. The Naïve Bayes classification is able to perform well with minimal computational effort and high speed.[41-42]

### 3.4.3. Neural network (NN)

A group of neurons that transfer electrical signal patterns is called an artificial neural network (ANN). The foundation of artificial neural networks (ANN) is biological neural networks, such those found in the human brain. These systems are particularly well-suited for resolving intricate issues that conventional approaches would find challenging [43-45].

ANN consists of input hidden and output layers:

Input layer: includes input units that represent all raw information that is provided to the network [46].

Hidden layer: consists of hidden units that are formed based on the behavior of input units and weighted neurons connected to the input.

Output layer: acts on the basis of hidden units and their characteristics and weight neurons.

### 4. Discussion and results

Comparison of algorithms:
SVM and: Random Forest
-Results suggest that Random Forest excels in complex data processing and runtime due to its set structure.

Table3. Summary of results

|  | Precision for normal | Recall for AD | Recall for normal |  | Classification accuracy | Precision for AD |
|---|---|---|---|---|---|---|
| MRMR-decision tree | 0.7987 | 0.7538 | 0.9071 |  | 83.33 | 0.8829 |
| CFS-decision | 0.7771 | 0.7154 | 0.9214 |  | 82.22 | 0.8942 |
| SDS-decision tree | 0.8313 | 0.7923 | 0.9500 |  | 87.41 | 0.9364 |
| MRMR-Naïve Bayes | 0.7692 | 0.7000 | 0.9286 |  | 81.85 | 0.901 |
| CFS-Naïve Bayes | 0.7711 | 0.7077 | 0.9143 |  | 81.48 | 0.8846 |
| SDS-Naïve Bayes | 0.8425 | 0.8077 | 0.9571 |  | 88.52 | 0.9459 |
| MRMR-neural network | 0.8221 | 0.7769 | 0.9571 |  | 87.04 | 0.9439 |
| CFS- neural network | 0.8012 | 0.7462 | 0.95 |  | 85.19 | 0.9327 |
| SDS- neural network | 0.859 | 0.8308 | 0.9571 |  | 89.63 | 0.9474 |

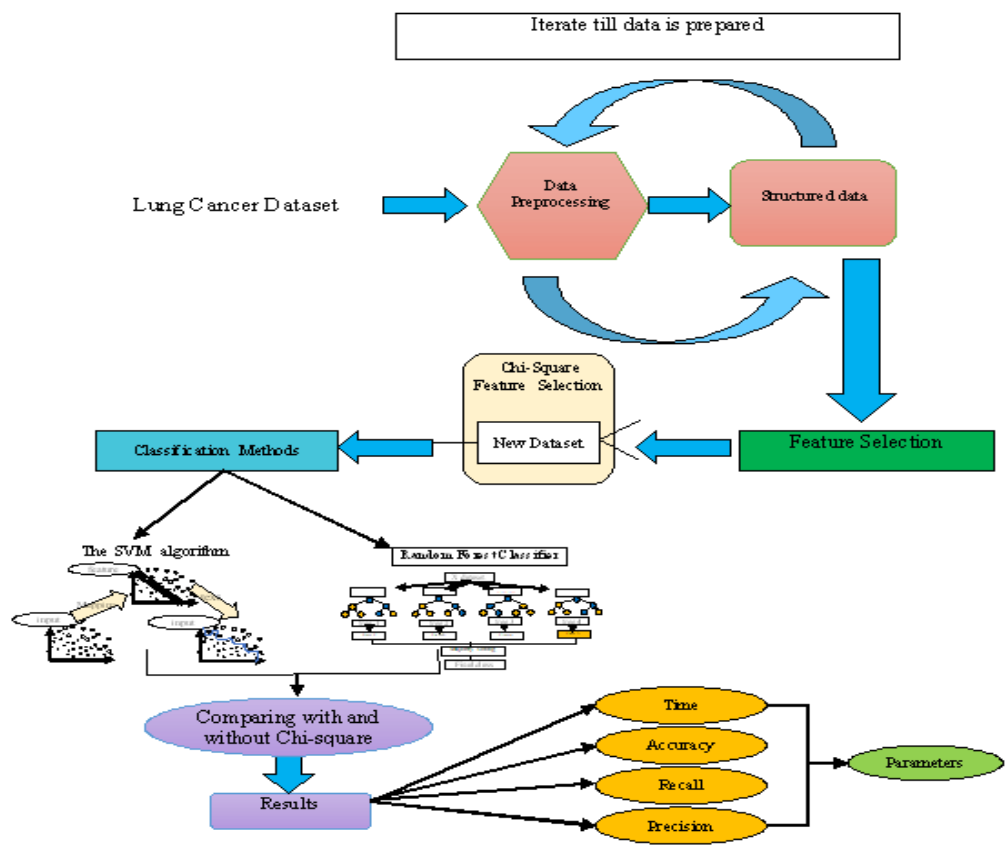

Figure 3: Diagram of the proposed

-SVM showed good ability to separate linear and nonlinear data using the appropriate kernel.

**The importance of feature selection:**

-Feature selection using statistical methods such as Chi-squared and advanced techniques such as SVM-RFE reduced computational complexity and increased the accuracy of models.

-These methods showed high efficiency, especially in high-dimensional microarray data.

Suggestions for future research:

-It is recommended that future research be done on larger and more diverse data to make the results more generalizable.

Combining statistical methods and machine learning can help improve the accuracy and efficiency of diagnosis.

Finally, this study highlights the important role of feature selection and advanced algorithms in improving the rapid and accurate diagnosis of lung cancer.

## 5. Conclusion

In order to increase the precision of lung cancer diagnosis, this study looked at a combination of feature selection techniques and machine learning models. According to the study's findings, the SVM model and the Chi-squared Test feature selection technique significantly improve lung cancer diagnostic accuracy. These results imply that combining the two techniques is a novel and successful strategy for lung cancer early detection.

However, this study also has limitations. Using a specific dataset and a limited number of models can affect the generalization of results. Also, the lack of examination of the impact of other clinical factors on diagnostic accuracy is another limitation of this study.

It is suggested that deeper learning models and bigger, more varied datasets be used in future studies. The effect of data balance on model performance and the creation of automated detection systems based on these models can also be investigated.

Lastly, the study's findings demonstrate the potential significance of machine learning in the early diagnosis of lung cancer. More research and the removal of current restrictions will enable the development of more precise and trustworthy diagnostic methods, which will benefit lung cancer patients' results during treatment.

**Acknowledgement**
The authors would like to thank all the organizations that provided data for this work.

**References**

[1] G. A. P. Singh and P. K. Gupta, "Performance analysis of various machine learning-based approaches for detection and classification of lung cancer in humans," *Neural Comput Appl*, vol. 31, no. 10, pp. 6863–6877, 2019.

[2] Duan, H., & Mirzaei, A. (2023). Adaptive Rate Maximization and Hierarchical Resource Management for Underlay Spectrum Sharing NOMA HetNets with Hybrid Power Supplies. Mobile Networks and Applications, 1-17.

[3] N. Banerjee and S. Das, "Prediction lung cancer–in machine learning perspective," in *2020 International conference on computer science, engineering and applications (ICCSEA)*, 2020, pp. 1–5.

[4] Mirzaei, A., & Najafi Souha, A. (2021). Towards optimal configuration in MEC Neural networks: deep learning-based optimal resource allocation. Wireless Personal Communications, 121(1), 221-243.

[5] S. H. Hawkins *et al.*, "Predicting outcomes of nonsmall cell lung cancer using CT image features," *IEEE access*, vol. 2, pp. 1418–1426, 2014.

[6] Somarin, A. M., Barari, M., & Zarrabi, H. (2018). Big data based self-optimization networking in next generation mobile networks. Wireless Personal Communications, 101(3), 1499-1518.

[7] . Alam, S. Alam, and A. Hossan, "Multi-stage lung cancer detection and prediction using multi-class svm classifie," in *2018 International conference on computer, communication, chemical, material and electronic engineering (IC4ME2)*, 2018, pp. 1–4

[8] Narimani, Y., Zeinali, E., & Mirzaei, A. (2022). QoS-aware resource allocation and fault tolerant operation in hybrid SDN using stochastic network calculus. Physical Communication, 53, 101709

[9] V. Jackins, S. Vimal, M. Kaliappan, and M. Y. Lee, "AI-based smart prediction of clinical disease using random forest classifier and Naive Bayes," *J Supercomput*, vol. 77, no. 5, pp. 5198–5219, 2021

[10] Mirzaei, A. (2022). A novel approach to QoS-aware resource allocation in NOMA cellular HetNets using multi-layer optimization. Concurrency and Computation: Practice and Experience, 34(21), e7068.

[11] K. Kancherla and S. Mukkamala, "Early lung cancer detection using nucleus segementation based features," in *2013 IEEE Symposium on Computational Intelligence in Bioinformatics and Computational Biology (CIBCB)*, 2013, pp. 91–95.

[12] Jahandideh, Y., & Mirzaei, A. (2021). Allocating Duplicate Copies for IoT Data in Cloud Computing based on Harmony Search Algorithm. IETE Journal of Research, 1-14.

[13] T. Viéville and S. Crahay, "Using an hebbian learning rule for multi-class svm classifiers," *J Comput Neurosci*, vol. 17, pp. 271–287, 2004.

[14] Mirzaei, A., Barari, M., & Zarrabi, H. (2019). Efficient resource management for non-orthogonal multiple access: A novel approach towards green hetnets. Intelligent Data Analysis, 23(2), 425-447.

[15] B. Schölkopf, A. Smola, and K.-R. Müller, "Nonlinear component analysis as a kernel eigenvalue problem," *Neural Comput*, vol. 10, no. 5, pp. 1299–1319, 1998.

[16] Mirzaei, A., & Rahimi, A. (2019). A Novel Approach for Cluster Self-Optimization Using Big Data Analytics. Information Systems & Telecommunication, 50.

[17] N. Cristianini and J. Shawe-Taylor, *An introduction to support vector machines and other kernel-based learning methods*. Cambridge university press, 2000.

[18] Rad, K. J., & Mirzaei, A. (2022). Hierarchical capacity management and load balancing for HetNets using multi-layer optimisation methods. International Journal of Ad Hoc and Ubiquitous Computing, 41(1), 44-57.

[19] Z. Rustam and A. S. Talita, "Fuzzy kernel k-medoids algorithm for multiclass multidimensional data classification," *J Theor Appl Inf Technol*, vol. 80, no. 1, pp. 147–151, 2015.

[20] Barari, M., Zarrabi, H., & Somarin, A. M. (2016). A New Scheme for Resource Allocation in Heterogeneous Wireless Networks based on Big Data. Bulletin de la Société Royale des Sciences de Liège, 85, 340-347

[21] Y. Tang, Y.-Q. Zhang, and Z. Huang, "Development of two-stage SVM-RFE gene selection strategy for microarray expression data analysis," *IEEE/ACM Trans Comput Biol Bioinform*, vol. 4, no. 3, pp. 365–381, 2007.

[22] Ziaeddini, A., Mohajer, A., Yousefi, D., Mirzaei, A., & Gonglee, S. (2022). An optimized multi-layer resource management in mobile edge computing networks: a joint computation offloading and caching solution. *arXiv preprint arXiv:2211.15487*.

[23] Mirzaei, A. (2021). QoS-aware Resource Allocation for Live Streaming in Edge-Clouds Aided HetNets Using Stochastic Network Calculus.

[24] K.-B. Duan, J. C. Rajapakse, H. Wang, and F. Azuaje, "Multiple SVM-RFE for gene selection in cancer classification with expression data," *IEEE Trans Nanobioscience*, vol. 4, no. 3, pp. 228–234, 2005.

[25] Mikaeilvand, N., Ojaroudi, M., & Ghadimi, N. (2015). Band-Notched Small Slot Antenna Based on Time-Domain Reflectometry Modeling for UWB Applications. The Applied Computational Electromagnetics Society Journal (ACES), 682-687.

[26] Hozouri, A., EffatParvar, M., Yousefi, D., & Mirzaei, A. Scheduling algorithm for bidirectional LPT.

[27] Y. Tang, Y.-Q. Zhang, Z. Huang, X. Hu, and Y. Zhao, "Recursive fuzzy granulation for gene subsets extraction and cancer classification," *IEEE Transactions on Information Technology in Biomedicine*, vol. 12, no. 6, pp. 723–730, 2008.

[28] Nemati, Z., Mohammadi, A., Bayat, A., & Mirzaei, A. (2024). Fraud Risk Prediction in Financial Statements through Comparative Analysis of Genetic Algorithm, Grey Wolf Optimization, and Particle Swarm Optimization. Iranian Journal of Finance, 8(1), 98-130.

[29] I. Guyon, J. Weston, S. Barnhill, and V. Vapnik, "Gene selection for cancer classification using support vector machines," *Mach Learn*, vol. 46, pp. 389–422, 2002.

[30] Zhang, S., Madadkhani, M., Shafieezadeh, M., & Mirzaei, A. (2019). A novel approach to optimize power consumption in orchard WSN: Efficient opportunistic routing. Wireless Personal Communications, 108(3), 1611-1634.

[31] Yousefi, D., Yari, H., Osouli, F., Ebrahimi, M., Esmalifalak, S., Johari, M., ... & Mirzapour, R. Energy Efficient Computation Offloading and Virtual Connection Control in. *learning (DL)*, *44*, 43.

[32] Z. Rustam and S. A. A. Kharis, "Comparison of support vector machine recursive feature elimination and kernel function as feature selection using support vector machine for lung cancer classification," in *Journal of Physics: Conference Series*, 2020, p. 12027.

[33] Nemati, Z., Mohammadi, A., Bayat, A., & Mirzaei, A. (2024). The impact of financial ratio reduction on supervised methods' ability to detect financial statement fraud. Karafan Quarterly Scientific Journal.

[34] R. Kohad and V. Ahire, "Application of machine learning techniques for the diagnosis of lung cancer with ANT colony optimization," *Int J Comput Appl*, vol. 113, no. 18, pp. 34–41, 2015.

[35] Nemati, Z., Mohammadi, A., Bayat, A., & Mirzaei, A. (2023). Financial Ratios and Efficient Classification Algorithms for Fraud Risk Detection in Financial Statements. International Journal of Industrial Mathematics.

[36] F. T. Johora, M. H. Jony, P. Khatun, and H. K. Rana, "Early Detection of Lung Cancer from CT Scan Images Using Binarization Technique," 2018.

[37] Nemati, Z., Mohammadi, A., Bayat, A., & Mirzaei, A. (2025). Fraud Prediction in Financial Statements through Comparative Analysis of Data Mining Methods. International Journal of Finance & Managerial Accounting, 10(38), 151-166.

[38] H. Alhakbani and M. M. Al-Rifaie, "Feature selection using stochastic diffusion search," in *Proceedings of the Genetic and Evolutionary Computation Conference*, 2017, pp. 385–392.

[39] Mirzaei, A., & Zandiyan, S. (2023). A Novel Approach for Establishing Connectivity in Partitioned Mobile Sensor Networks using Beamforming Techniques. arXiv preprint arXiv:2308.04797.

[40] S. D. Jadhav and H. P. Channe, "Comparative study of K-NN, naive Bayes and decision tree classification techniques," *International Journal of Science and Research (IJSR)*, vol. 5, no. 1, pp. 1842–1845, 2016.

[41] Allahviranloo, T., & Mikaeilvand, N. (2011). Non zero solutions of the fully fuzzy linear systems. Appl. Comput. Math, 10(2), 271-282.

[42] Nematia, Z., Mohammadia, A., Bayata, A., & Mirzaeib, A. (2024). Predicting fraud in financial statements using supervised methods: An analytical comparison. International Journal of Nonlinear Analysis and Applications, 15(8), 259-272.

[43] S. Senthil and B. Ayshwarya, "Lung cancer prediction using feed forward back propagation neural networks with optimal features," *International Journal of Applied Engineering Research*, vol. 13, no. 1, pp. 318–325, 2018.

[44] Nemati, Z., Mohammadi, A., Bayat, A., & Mirzaei, A. (2024). Metaheuristic and Data Mining Algorithms-based Feature Selection Approach for Anomaly Detection. IETE Journal of Research, 1-15.

[45] Nematollahi, M., Ghaffari, A., & Mirzaei, A. (2024). Task and resource allocation in the internet of things based on an improved version of the moth-flame optimization algorithm. Cluster Computing, 27(2), 1775-1797.

[46] Yousefi, D., Farhad, F., Abed, M., & Gavidel, S. (2024). Presenting a new approach in security in inter-vehicle networks (VANET). *arXiv preprint arXiv:2411.19002*.